%% file: main.tex
\documentclass{article}
\usepackage{mathtools}
\usepackage{bbm}
\usepackage{dsfont}
\usepackage{verbatim}
\usepackage{comment}
\usepackage[ruled,vlined]{algorithm2e}
\usepackage{algorithm,algorithmic}

\usepackage[dvipdfm]{graphicx} 
\usepackage{bmpsize}

\usepackage{microtype}
\usepackage{subfigure}
\usepackage{booktabs} 
\usepackage{hyperref}
\usepackage{wrapfig}

\usepackage[preprint, nonatbib]{neurips_2019}

\usepackage{tikz}
\usetikzlibrary{intersections,shapes.arrows}

\usepackage{amsthm, amsmath, amssymb, amsfonts}

\DeclareMathOperator*{\argmax}{arg\,max}

\usepackage{comment}
\usepackage{mathrsfs}
\usepackage{enumitem}
\usepackage{bbm}

\title{Ready Policy One: World Building Through Active Learning}

\author{
	Philip Ball$^*$\\
	University of Oxford \\
	\texttt{ball@robots.ox.ac.uk} \\
	 \And
	 Jack Parker-Holder$^*$\\
	 University of Oxford \\
	 \texttt{jackph@robots.ox.ac.uk} \\
	 \And
	 Aldo Pacchiano \\
	 UC Berkeley \\
	 \texttt{pacchiano@berkeley.edu} \\
	 \AND
	 Krzysztof Choromanski \\
	 Google Brain Robotics \\
	 \texttt{kchoro@google.com} \\
	 \And
	 Stephen Roberts \\
	 University of Oxford \\
	 \texttt{sjrob@robots.ox.ac.uk} \\
	 \\[-50.0ex]
   }

\begin{document}
\maketitle

\def\thefootnote{*}\footnotetext{Equal contribution.}

\begin{abstract}
Model-Based Reinforcement Learning (MBRL) offers a promising direction for sample efficient learning, often achieving state of the art results for continuous control tasks. However many existing MBRL methods rely on combining greedy policies with exploration heuristics, and even those which utilize principled exploration bonuses construct dual objectives in an ad hoc fashion. In this paper we introduce Ready Policy One (RP1), a framework that views MBRL as an active learning problem, where we aim to improve the world model in the fewest samples possible. RP1 achieves this by utilizing a hybrid objective function, which crucially adapts during optimization, allowing the algorithm to trade off reward v.s. exploration at different stages of learning. In addition, we introduce a principled mechanism to terminate sample collection once we have a rich enough trajectory batch to improve the model. We rigorously evaluate our method on a variety of continuous control tasks, and demonstrate statistically significant gains over existing approaches.
\end{abstract}

\input{intro.tex}
\input{related.tex}

\input{preliminaries.tex}
\input{readypolicyone.tex}

\input{theory.tex}

\input{experiments.tex}

\input{conclusions.tex}

\bibliographystyle{abbrv}
\bibliography{refs}

\input{appendix.tex}

\end{document}

%% file: intro.tex
\section{Introduction}

Reinforcement Learning (RL) considers the problem of an agent learning to construct of actions that result in an agent receiving high rewards in a given environment. This can be achieved in various ways such as: learning an explicit mapping (\textit{a policy}) from states to actions that maximizes expected return (policy gradients), or inferring such a mapping by calculating the expected return for a given state-action pair (TD-control methods). 
Model-Based Reinforcement Learning (MBRL) seeks to improve the above by learning a model of the dynamics (from agent's interactions with the environment) that can be leveraged across many different tasks (transferability) and for planning, which is substantially less expensive than a real environment (which plays crucial role in robotics applications). 

A series of recent work \cite{worldmodels, planet, ME-TRPO, mbmpo, pets, simple, nuzero} illustrate the benefits of MBRL-approaches that allow us to decouple learning task-dependent policy and task-agnostic dynamics. With recent advances, MBRL approaches often outperform model-free methods \cite{benchmarkingmbrl}. However, these results are often overly sensitive to heuristics. 

In particular, many of these methods lack a principled mechanism to acquire data for training the model. This issue is circumvented in \cite{worldmodels}, since they only consider environments which can be explored with random policies. Other model-based approaches, such as \cite{ME-TRPO, modelfreehybrid,hafner2020dream}, rely on stochastic policies to aid exploration, and inevitably acquire redundant data which reduces sample efficiency. Such issues have been highlighted previously \cite{Schmidhuber1991CuriousMC, plantobesurprised}, and motivate the design of our algorithm. Concretely, we reduce the cost incurred from data collection by using active learning methods, and introduce an early stopping mechanism to address the issue of redundancy.

Efficient exploration is a challenge for existing RL algorithms, and is a core focus in model-free RL \cite{intrinsicRL, vime, rmax, lopes2012}. 
Despite often considering a principled objective, these methods generally contain a fixed temperature parameter, thus requiring hand engineering to determine the optimal degree of exploration. Our approach adjusts this parameter in an online manner from the collected trajectories, and we provide an information theoretic motivation for our exploration objective.

In this paper, we introduce a novel approach to acquiring data for training world models through exploration. Our algorithm, Ready Policy One (RP1), includes principled mechanisms which acquire data for model-based RL through the lens of Online Active Learning. Crucially, we continue to jointly optimize our policies for both reward and model uncertainty reduction, since we wish to avoid focusing on stochastic or challenging regions of the state space which have no impact on the task at hand. Therefore policies used for data collection also perform well in the true environment, and means we can use these policies for evaluation. Consequently, a separate `exploit' agent does not need to be trained \cite{explicitexploreexploit}.

To summarize, our key contributions are as follows:
\begin{itemize}
    \item Inspired by Active Learning, we train policies in a learned world model with the objective of acquiring data that most likely leads to subsequent improvement in the model.
    \item We introduce a novel early-stopping criteria for real-environment samples, reducing redundancy in expensive data collection.
    \item We adapt the objective function as learning progresses using an Online Learning mechanism.
\end{itemize}

The paper is structured as follows. In Section \ref{sec:related} we discuss related work. In Section \ref{sec:background} we describe our RL setting and introduce basic concepts. In Section \ref{sec:RP1} 
we introduce our method and related theory. Finally we demonstrate the effectiveness of our approach across a variety of continuous control tasks in Section \ref{sec:experiments} before concluding in Section \ref{sec:conclusion}, where we also mention some exciting future work.

%% file: related.tex
\section{Related Work}
\label{sec:related}

The Dyna algorithm \cite{dyna} is a canonical approach for model based reinforcement learning (MBRL), based on the principle of `trying things in your head', using an internal
model of the world. In its original form, Dyna contained an exploration bonus for each state-action pair, proportional to this uncertainty measure \cite{Sutton90integratedarchitectures}. A decade later, principled approaches were proposed for exploration \cite{rmax}, yet their guarantees were only possible in discrete settings with a finite number of states. 

In recent times there has been great deal of progress in MBRL, with success in Atari games \cite{simple, nuzero}, and dexterous manipulation \cite{nagab2019deep}, while progress has been made on continuous control benchmarks \cite{ME-TRPO, pets, mbmpo, whentotrust}. We note that our approach is orthogonal to these in the following ways: 1) our methods, or some subset of the methods we introduce, can be incorporated into existing MBRL algorithms; 2) we adhere to a strict Dyna style framework, and our methods are aimed at incorporating active learning into this paradigm. Other recent work \cite{worldmodels} shows that Dyna can be extended to latent-state dynamics models. We note that our framework could be extended to this setting, however we instead focus on efficient data acquisition through active learning.

Active approaches (i.e., acquiring data in a principled manner) in MBRL have been considered previously, for example in \cite{shyam,pathak,explicitexploreexploit}. Usually, ensembles of models are maintained, and an intrinsic reward, defined as some difference measure (i.e., KL-divergence, total variation) across the output of different models in the ensemble drives exploration. Such exploration might be ineffective however, as the policy may visit regions of the state space which have no relation to solving the task. This may also lead to unsafe exploration if deployed on a real robot. \cite{explicitexploreexploit} bears similarities to our work in that it aims to improve model generalization through exploration, and has a criteria to trade-off exploration and exploitation. However their approach to exploration is purely novelty-seeking, and they collect data until they discover a model (or subset of models) that can fully model the MDP in question (i.e., when novelty falls below a predefined value). Once a model is discovered, they implement a policy (through search) which exploits it to maximize performance. This has drawbacks mentioned above concerning wasted exploration in task-irrelevant states as well as unsafe exploration.

Also similar to our work is \cite{JapanPaper}, who explicitly use an active learning approach for solving an RL task; a robot arm hitting a ball. 
Our work differs in three significant ways: 1) our policies are trained inside a model, not on the true environment; 2) we actively limit the number of trajectories per collection phase based on the data (they introduce a heuristic); 3) we do not have access to an off-line measure of generalization, and therefore must introduce an online-learning mechanism to seek out the optimal setting of the exploration parameter.

Approaches that produce Bayes-optimal exploration and exploitation with a model \cite{PILCO, activeinference} are also of relevance, however these methodologies do not scale well to high dimensional tasks \cite{benchmarkingmbrl}.

Efficient exploration in environments with very sparse rewards also represents a relevant area of research. In such settings an agent follows its \textit{curiosity}, quantified by either: 1) rewarding areas of the state-space that reduce uncertainty in some internal model (i.e., inverse or forward dynamics models) \cite{intrinsicRL, mohamed2017, vime, pathak_agrawal, burda}; 2) rewarding un-visited areas of the state-space \cite{rmax,lopes2012,countbased}. Our approach to exploration leverages a model ensemble, and sits in the former category.

There has also been a great deal of work on using maximum entropy principles as a means for exploration in model-free RL \cite{sac}. The aim is to find rewarding behavior whilst maximizing some measure of entropy. We differ from these works in both what entropy is maximized (action entropy v.s. model prediction entropy) and by not having task-specific, fixed temperature parameter that trades off reward and entropy/surprise. In later maximum entropy work, temperature selection has been formulated as a constrained optimization problem, such that performance is maximized subject to some minimum level of policy entropy \cite{sac-v2}. In contrast, we select this parameter in an online manner that optimally improves our internal models.

%% file: preliminaries.tex
\section{Background}
\label{sec:background}

\subsection{RL Policies \& Markov Decision Processes}

A Markov Decision Process ($\mathrm{MDP}$, \cite{bellmanmdp}) is a tuple $(\mathcal{S},\mathcal{A},\mathrm{P},\mathrm{R})$. Here $\mathcal{S}$ and $\mathcal{A}$ stand for the sets of states and actions respectively, such that for $s_t, s_{t+1} \in \mathcal{S}$ and $a_t \in \mathcal{A}$: $\mathrm{P}(s_{t+1}| s_t, a_t)$ is the probability that the system/agent transitions from $s_t$ to $s_{t+1}$ given action $a_t$ and $\mathrm{R}(a_t,s_t,s_{t+1})$ is a reward obtained by an agent transitioning from $s_t$ to $s_{t+1}$ via $a_t$.

A policy $\pi_{\theta}:\mathcal{S} \rightarrow \mathcal{A}$ is a (possibly randomized) mapping (parameterized by $\theta \in \mathbb{R}^{d}$, e.g. weight of the neural network) from $\mathcal{S}$ to $\mathcal{A}$. Policy learning is the task of optimizing parameters $\theta$ of $\pi_{\theta}$ such that an agent applying it in the environment given by a fixed $\mathrm{MDP}$ maximizes total (expected/discounted) reward over given horizon $H$. In this paper we consider MDPs with finite horizons.

In most practical applications the $\mathrm{MDP}$ is not known to the learner. In MBRL, we seek to use a dataset $\mathcal{D} = \{(s_t, a_t), s_{t+1}\}_{t=1}^N$ of observed transitions to train a dynamics/world model $\hat{f}_{\phi}$ parameterized by $\phi$ to approximate the true dynamics function $f(s_{t+1} | s_t, a_t)$ such that $\hat{f}_{\phi}(s_{t+1}| s_t, a_t) \approxeq f(s_{t+1} | s_t, a_t)$.

We aim to construct rich $\mathcal{D}$s for learning accurate enough models $\hat{f}_{\theta}$, but only in those regions that are critical for training performant policies. 


\subsection{Sequential Model Based Optimization}

Consider a black box function $F:\mathbb{R}^{d} \rightarrow \mathbb{R}$ over some domain $\mathcal{X}$, whereby the goal is to find $x^* \in \mathcal{X}$ such that 
\begin{equation} 
    x^* = \underset{x \in \mathcal{X}}{\argmax} F(x) 
\end{equation} 
Sequential Model Based Optimization (SMBO, \cite{SMBO}) is a model-based black box optimization method which seeks to learn a surrogate model $\hat{F}$, within the true model $F$. Using the surrogate model, it is possible to determine which data should be collected to discover the optimum point of the real black box function $F$. The surrogate model is sequentially updated with the data collected in order to obtain better estimates of the true $F$, and this process is repeated until convergence or limited to a set number of iterations. 

Many MBRL algorithms follow this regime, by trying to model a true black box function $F$ using a world model $\hat{f}_\phi$ parameterized by $\phi$ as follows:
\begin{align}
    F(\theta) \approxeq& \sum_{t=0}^H  \gamma_t R(s_t, \pi_\theta(s_t), \hat{f}_\phi(\pi_\theta(s_t), s_t))
\end{align}where $\gamma_t \in [0,1] $ is a discount factor, actions are $\pi_\theta(s_t) = a_{t}$, and next states are generated by $s_{t+1} = \hat{f}_\phi(\pi_\theta(s_t), s_t)$. Subsequently, they seek to find a policy $\pi$ parameterized by $\theta$ that maximizes the acquisition function, which is just  $\text{max} F(\theta)$, then they will generate rollouts in the environment by using this policy and adding noise to the actions, such as via a stochastic policy.

Taking this view, essentially the majority of existing MBRL algorithms are conducting a form of SMBO with a greedy acquisition function. Any exploration conducted is simply a post hoc addition of noise in the hope of injecting sufficient stochasticity, but there is no principled mechanism by which the model may escape local minima.

Our approach is markedly different. We seek to conduct \textit{Active Learning}, whereby the goal is not just to find an optimal solution to $F$, but to learn an optimal surrogate model $\hat{F}^*$ through the world model $\hat{f}_{\phi^*}$. To do this, we train policies to maximize an acquisition function that trades-off reward and information, rather than just greedily maximizing reward. Given how the acquisition function in SMBO is an important component for finding optimal performance \cite{Jones1998EfficientGO}, it therefore makes sense to carefully design how we acquire data in the MBRL setting.

\subsection{Active Learning}
 
Active Learning considers the problem of choosing new data $\mathcal{D}' = \{x_i', y_i'\}_{i=1}^m$, which is a subset of a larger dataset i.e., $\mathcal{D}' \in \mathcal{D} = \{x_i, y_i\}_{i=1}^n$, such that a model $\mathcal{M}$ 
is most improved.

In traditional supervised learning, we usually have access to the entirety of $\mathcal{D}$ for training, but in the active learning setting we only have access to its feature vectors $\{x_i\}_{i=1}^n$, and need to query an oracle to obtain the corresponding labels $\{y_i\}_{i=1}^n$ which incurs a cost. Active learning aims to reduce this cost by iterating through the set of unlabeled feature vectors $\{x_i\}_{i=1}^n$, and determining which subset $\{x_i'\}_{i=1}^m$ would produce the greatest improvement in the performance of $\mathcal{M}$ should we train our models on the subset $\mathcal{D'} = \{x_i', y_i'\}_{i=1}^m$. The ultimate goal is to achieve the highest performance with the fewest number of queries (i.e., at the lowest cost).

In reinforcement learning, there is a key difference to the standard active learning setting; we do not have direct access to a dataset $\mathcal{D}$, and must instead generate $\mathcal{D}$ through placing a `sampling' policy $\pi_s$ in the environment, producing trajectories. We then assess these trajectories generated by $\pi_s$ and determine how to best train this policy to obtain trajectories in some oracle/true environment that will benefit performance. Framing the problem this way is done in \cite{JapanPaper}, where it is cheap to generate and evaluate trajectories, but expensive to obtain labels.

In the Dyna-style approach \cite{dyna} that we adopt, this means training our sampling policy $\pi_s$ in the world model exclusively, then using this policy to collect samples in the real world. Through the lens of active learning, it is therefore important to train $\pi_s$ in our world model such that the resultant trajectories maximize our world model performance, therefore maximizing our final policy performance. This is because it is free, from a data collection perspective, to train in the world model, but acquiring data in the real environment incurs an expense.

Concretely, we wish to train a policy that performs well in the real environment in as few samples as possible, and we identify that having a robust and generalizable world model as being paramount to achieving this. We observe that state-action pairs which cause high uncertainty/disagreement in the world model are likely to be parts of the state space that our world model is poor at modeling (as observed in \cite{ObsDropout}, where solving `difficult' environment dynamics is important for learning optimal policies).

However we cannot simply target regions of disagreement; we are primarily concerned with maximizing policy performance, so wish to explore regions of disagreement that are also critical to solving the task. This becomes a classic explore/exploit dilemma; how much do we want our sampling policy to explore (i.e., find states with poor generalization) or exploit (i.e., visit known high value states) when acquiring new real samples. In order to manage this trade-off, we leverage an online active learning approach similar to \cite{OnlineActiveLearning}, where model generalization feedback from the gathered trajectories can be used to determine the degree to which we explore or exploit in the future.

\subsection{Online Learning}

Online Learning is a family of methods that is used when a learner tackles a decision-making task by taking actions $a \in \mathcal{A}$, whilst learning from a sequence of data $z_1, z_2, \dots, z_N$ that arrive in incremental rounds $n$. The learner must take an action at the beginning of each round, and the aim is to minimize cumulative regret, which is usually defined as the difference in some loss $\ell(a, z_n)$ between the actual action taken and the optimal action that could have been taken at that round:
\begin{align}
\sum_{n=0}^N \left( \ell(a_n, z_n) - \min_{a\in\mathcal{A}}\ell(a, z_n) \right) .\notag
\end{align} 
At each round $n$ the learner does not know what the consequence of some action $a$ will be, and only receives feedback after submitting its chosen action $a_n$. Based on this feedback the learner then updates how it selects actions in the future. In our approach we consider the set of actions $\mathcal{A}$ to be the degree to which our policy explores, and the loss $\ell$ to be a normalized generalization error from the data collected. The aim is to therefore ensure that generalization error (i.e., RMSE on the new data) is maximized at each round. Since the task of maximizing generalization is both noisy and stochastic (i.e., the optimal degree of exploration may vary as we collect data), careful design of this algorithm is required. 

%% file: readypolicyone.tex
\section{Ready Policy One}
\label{sec:RP1}

Here we introduce our main algorithm, \textit{Ready Policy One} (RP1). The key differences between RP1 and existing state of the art MBRL methods are as follows:
\begin{enumerate}
    \item By taking an Active Learning approach rather than focusing on greedy optimization, RP1 seeks to directly learn the best \textit{model}, rather than learning the best \textit{policy}, and indirectly learning the best model to achieve this objective. 
    \item We introduce a principled Online Learning-inspired framework, allowing RP1 to adapt the level of exploration in order to optimally improve the model in the fewest number of samples possible. 
    \item We introduce a mechanism to stop gathering new samples in any given collection phase when the incoming data resembles what we have already acquired during that phase.
\end{enumerate}

The algorithm begins in a similar fashion to other MBRL methods, by sampling initial transitions with a randomly initialized policy. In the Dyna framework, a policy is then trained inside $\hat{f}_\phi$, and then subsequently used to gather new data. Typically, random noise is added to the policy to induce exploration. Other methods consider a hybrid objective. In RP1, consider training a \textit{sampling policy}, parameterized by $\theta_t$, to optimize the following objective:
\begin{align}
    \pi_{\theta_t} = \text{max}[\mathbb{E}_{\tau \sim \pi_{\theta_t}} [(1-\lambda)\mathcal{R}(\tau) +  \lambda\sigma(\mathcal{R}(\tau)]]
\label{eqn:explore}
\end{align}
where $\mathcal{R}(\tau) = \sum_{i=0}^H r_i$, $r_{i}=R(s_{i},a_{i},s_{i+1})$ and $\sigma(\mathcal{R}(\tau)) = \sum_{i=1}^H \sqrt{\frac{\sum_{j=1}^M (r_i^j - \bar{r_i})^2}{M-1}}$. This $\lambda$ value is chosen before training the policy in the model, and is selected using an online learning algorithm mechanism detailed in \ref{subsec:online-learning}. Hence $\lambda$ defines the relative weighting of reward and reward variance, with $\lambda = 0$ training a policy that only maximizes expected return (model-guided exploitation)\footnote{$\lambda = 0$ corresponds to the same objective as in prior MBRL work, such as \cite{ME-TRPO}.}, and $\lambda = 1$ training a policy that only maximizes variance/disagreement per time step (model-guided exploration). In reality we limit $\lambda$ to $[0.0, 0.5]$ as we wish any exploration to be guided by the reward signal. As a consequence, there is no need to train a separate `exploit' policy, since we find policies trained in this way provide significant gains over commonly used approaches. This is mirrors the fact that MaxEnt strategies obtain high-performance in deterministic environments \cite{eysenbach2019maxent}.

\subsection{Information Gain in the Model as Maximum Entropy}

Our objective in Equation \ref{eqn:explore} is inspired by curiosity driven exploration via model disagreement \cite{pathak}. When shown a new unlabeled feature $x'$ comprising a state $s \in \mathcal{S}$ and action $a \in \mathcal{A}$ pair, the model has an existing `prior' $p(r | x', \mathcal{D})$ where $r$ is the predicted reward and $\mathcal{D}$ is the data seen so far. After obtaining the label $s'$ we have a new datapoint $D'$ (i.e., $D' = (s,a,s')$), and can update the world model, producing a `posterior' $p(r|s,a,\{\mathcal{D} \cup D' \})$. In our case, reward is a deterministic function of this triple (i.e., $r = R(D')$, see Appendix \ref{sec:implementation} for details). As such, we define the \textit{Information Gain} (IG) in the reward as the KL divergence between the model posterior after observing $D'$ and its' respective prior at $x'$, as follows:
\begin{align}
    \mathcal{IG}(r;x') = D_{\mathrm{KL}}[p(r|x',\{\mathcal{D} \cup D' \})|| p(r|x',\mathcal{D})]] \\
    = \int p(r|x',\{\mathcal{D} \cup D' \}) \log \frac{p(r|x',\{\mathcal{D} \cup D' \})}{p(r|x',\mathcal{D})} \mathrm{d}r.
\end{align}
We observe that to maximize $\mathcal{IG}(r;x')$, we must sample $x'$ appropriately, which is our only degree freedom (through policy training). Because we cannot tractably calculate this quantity for all $D'$, one approach to maximize information gain is to ensure that the prior assigns low mass to all regions of $r$ (i.e., minimize the denominator over all $r$). In order to do this we would select the improper prior over the continuous variable $r$\footnote{This follows the proposal in \cite{itila} when selecting priors in the face of uncertainty.}.
In our setting however, the model takes the form of an empirical Gaussian distribution, formed from the sample mean and variance of the individual models in the ensemble. Therefore we would like to like $p(r|x',\mathcal{D})$ such that the following is minimized:
\begin{align}\label{eqn:klimproper}
    D_{\mathrm{KL}}[p(r|x',\mathcal{D})|p_0(r)]
\end{align}
where $p_0(r)$ is an improper prior. The only way that Equation \ref{eqn:klimproper} is minimized is when the differential entropy of $p(r|x',\mathcal{D})$ is maximized. The differential entropy of a Gaussian is well known, given by $h(x) = \ln(\sqrt{2\pi}\sigma) + \frac{1}{2}$. Therefore to maximize the entropy of $p(r|x',\mathcal{D})$, and maximize information gain in the face of uncertainty, we need to maximize its variance. This is achieved by training policies that generate trajectory tuples $x'=(s,a)$ which cause high variance in the model, and is analogous to information based acquisition functions in active learning \cite{mackay1992}.

This motivates the second term of Equation \ref{eqn:explore}, where we show the objective function for our exploration policies. Essentially, we are seeking to maximize information gain \textit{in the model} through maximizing model entropy over the reward. This is in contrast to other maximum entropy approaches, which seek to maximize entropy over the action distribution \cite{sac}, aiming to succeed at the task while acting as randomly as possible.

It is also possible to maximize information gain for next state predictions (as opposed to rewards), and this is similar to the approach in \cite{pathak}. However in practice we find that maximizing reward variance results in better performance (see Section \ref{sec:experiments}).

\subsection{Online Learning Mechanism}\label{subsec:online-learning}

We use the Exponential Weights framework to derive an adaptive algorithm for the selection of $\lambda$. In this setup we consider $k$ experts making recommendations at the beginning of each round. After sampling a decision $i_t \in \{ 1, \cdots, k\}$ from a distribution $\mathbf{p}^t \in \Delta_k$ with the form $\mathbf{p}^t(i) \propto \exp\left(  \ell_t(i)   \right)$ the learner experiences a loss $l_{i_t}^t \in \mathbb{R}$. The distribution $\mathbf{p}^t$ is updated by updating $\ell_t$ as follows:
\begin{equation}\label{equation::exponential_weights_update}
    \ell_{t+1}(i) = \begin{cases}
        \ell_t(i) + \eta \frac{l_i^t}{\mathbf{p}^t(i)} & \text{if } i = i_t \\
        \ell_t(i) &\text{o.w. }
        \end{cases}
\end{equation}
For some step size parameter $\eta$. 

In our case we consider the case when the selection of $\lambda_t$ is thought as choosing among $k$ experts which we identify as the different values $\{\lambda_i \}_{i=1}^k$. The loss we consider is of the form $l_{i_t} = \hat{G}_{\phi_{t}}(\theta_{t+1})$, where ${G}_{\phi_{t}}(\theta_{t+1})$ is the RMSE of the model under parameters $\phi_t$ on data collected using $\pi_{\theta_{t+1}}$, and $\theta_{t+1}$ is the parameter of the policy trained under the choice $\lambda_{i_t}$, after incorporating into the model the data collected using the previous policy $\pi_{\theta_t}$. We then perform a normalization of $G$ (see Appendix \ref{sec:implementation} for details), hence $\hat{G}$. Henceforth we denote by $\mathbf{p}^t_\lambda$ the exponential weights distribution over $\lambda$ values at time $t$. 


\vspace{-3mm}
{\centering
\begin{minipage}{.99\linewidth}
    \centering\begin{algorithm}[H]
    \textbf{Input:} step size $\eta$, number of timesteps $T$. \\
    \textbf{Initialize:} $\mathbf{p}_\lambda^1$ as a uniform distribution. \; \\
    \For{$t= 1, \ldots, T-1$}{
      1. Select $i_t \sim \mathbf{p}_\lambda^t$ and $\lambda_t = \lambda_{i_t}$. \;\\
      2. Use  Equation \ref{equation::exponential_weights_update} to update  $\mathbf{p}_\lambda^t$ with\\
      \vspace{-2mm}
      \begin{equation*}
          l_{i_t}^t = \hat{G}_{\phi_t}(\theta_{t+1})
      \end{equation*}
      \vspace{-4mm}
     }
     \caption{Online Learning Mechanism}
    \label{Alg:ol}
    \end{algorithm}
\end{minipage}
}

Our version of Exponential Weights algorithm also known as Exponentially Weighted Average Forecaster \cite{cesa2006prediction} is outlined in Algorithm \ref{Alg:ol}.

In practice and in order to promote more effective exploration over $\lambda$ values we sample from a mixture distribution where $\mathbf{p}_\lambda^t$ is not proportional to $\exp\left(  \ell_t   \right)$ but it is a mixture between this exponential weights distribution and the uniform over $[k]$. In other words, let $\epsilon >0$ be a small parameter. With probability $1-\epsilon$ the produce $i_t$ as a sample from an exponential weights distribution proportional to $\exp\left(  \ell_t   \right)$, and with probability $\epsilon$ it equals a uniform index from $1, \cdots, k$.

\subsection{Diverse Sample Collection}

Consider the problem of online data acquisition from a policy in an environment. At each timestep we receive a set of datapoints $\{x_1, \dots, x_H\} \sim \pi_\theta$ corresponding to the concatenation of each state and action in a trajectory. At timestep $t$ we have a dataset $X_t = \{x_1, \dots, x_n\} \subset \mathbb{R}^d$, where $X_t \in \mathbb{R}^{d \times n}$ sampled from the sampling policy. We represent this data in the form of the Singular Value Decomposition $(\mathrm{SVD})$ of the symmetric matrix, $\mathrm{Cov}_{t} = \frac{1}{n} X_t X_t^\top =  \mathbf{Q}_{t}^{\top}\Sigma_{t}\mathbf{Q}_{t} \in \mathbf{R}^{d}$.

Equipped with this representation, we take the top $k$ eigenvalues $\lambda_{i}$ of $\mathrm{Cov}_{t}$, where $k$ is smallest such that: $\sum_{i=1}^{k}\lambda_{i} \geq (1-\delta) \sum_{i=1}^{d} \lambda_{i}$ for some parameter $\delta > 0$, and take $n_{t}=k$. Next we take the corresponding eigenvectors $\mathbf{u}_{1},...,\mathbf{u}_{k} \in \mathbb{R}^{d}$ and let $\mathbf{U} \in \mathbb{R}^{d \times k}$ be obtained by stacking them together. We define the Active Subspace \cite{constantineAS} $\mathbf{U}^{\mathrm{act}} \in \mathbb{R}^{d \times k}$ as $\mathcal{L}_{\mathrm{active}}   \overset{\mathrm{def}}{=} \mathrm{span}\{\mathbf{u}_{1},...,\mathbf{u}_{k}\}$. $\mathbf{U}^{\mathrm{act}}$ is an orthonormal basis of $\mathcal{L}_{\mathrm{active}}$.

We use $\mathbf{U}^{\mathrm{act}}$ to evaluate new data. After we collect $n'$ new samples $V_{t+1} \in \mathbb{R}^{ d\times n'}$, we form a covariance matrix with this new data as $\frac{1}{n'}V_{t+1}V_{t+1}^T \in \mathbb{R}^{d \times d} $ and project it onto $\mathbf{U}^{\mathrm{act}}$. We define the residual at timestep $t$, $r_t$, as follows:
\begin{equation}\label{eqn:residual}
    r_t =\frac{ \mathrm{tr}(V_{t+1} V_{t+1}^T - \mathrm{U}\mathrm{U}^T V_{t+1} V_{t+1}^T \mathrm{U}\mathrm{U}^T ) }{\mathrm{tr}(V_{t+1} V_{t+1}^T) }
\end{equation}
Where $\mathrm{tr}$ denotes the trace operator. After evaluating $r_t$ we append the new data $V_{t+1}$ to $X_t$ to form $X_{t+1} \in \mathbb{R}^{ d \times (n+n')}$. Intuitively, $r_t$ tells us how much of the new data could not be explained by the principal components in the data collected thus far. We stop collecting data and proceed to retrain the model once $r_t < \alpha$, where $\alpha$ is a proxy for $\delta$. The full procedure is presented in Algorithm \ref{Alg:earlystop}. 

Let $q_t$ be the probability that at timestep $t$, step $4.$ of Algorithm \ref{Alg:earlystop} is executed (i.e., $q_t = \mathrm{P}(r_t < \alpha)$). The evolution of $q_t$ operates in roughly two phases. First, the algorithm tries to collect data to form an accurate estimate of the covariance matrix $\mathrm{Cov}_{t}$ and a stable estimator $\mathbf{U}^{\mathrm{act}}$. During this phase, $q_t$ is small as it is roughly the probability of two random samples $X, X' \in  \mathbb{R}^{d \times n'}$ aligning. After $t_0$ steps when the algorithm has built a stable estimator of $\mathbf{U}^{\mathrm{act}}$, the stopping probability stabilizes to a value $q^*$ that solely depends on the trajectories intrinsic noise. Both the length of $t_0$ and  the magnitude of $q^*$ scale with the trajectories' noise. If the trajectories have little noise both $t_0$ and $q^*$ are small. On the other hand, if the trajectories have high noise, the early stopping mechanism will take longer to trigger.

{\centering
\scalebox{0.95}{\begin{minipage}{1.05\linewidth}
    \begin{algorithm}[H]
    \textbf{Input:} thresholds $\alpha, \delta$, maximum number of samples $T$. \\
    \textbf{Initialize:} training set $X = \emptyset$ \; \\
    Collect initial samples $X_1 = \{x_1, \dots, x_H\}$. \\
    \For{$t= 2, \ldots, T-1$}{
      1. Compute Active Subspace $\mathbf{U}^{\mathrm{act}}$ as the result of stacking together some orthonormal basis of $\mathcal{L}_{\mathrm{active}}   \overset{\mathrm{def}}{=} \mathrm{span}\{\mathbf{u}_{1},...,\mathbf{u}_{r}\}$ where the vectors $\mathbf{u}_{i}$ correspond to the top $k$ eigenvalues of the covariance matrix $\mathrm{Cov}_t$. \;\\
      2. Produce samples $V_{t+1}$ via the sampling policy \;\\
      3. Calculate the residual $r_t$ using Equation \ref{eqn:residual}. \;\\
      4. Stop collecting data if $r_t<\alpha$
     }
     \caption{Early Stopping Mechanism}
    \label{Alg:earlystop}
    \end{algorithm}
\end{minipage}}
}

This dynamic approach to determining the effective `batch size' of the incoming labeled data is similar to \cite{dynamicbatchsizeactive}, whereby feedback from unsupervised learning is used to control the amount of data collected per batch. However, we do this in a more instantaneous fashion, leveraging data collected so far to determine when to stop.

\subsection{The Algorithm}

{\centering
\begin{minipage}{.99\linewidth}
    \begin{algorithm}[H]
    \textbf{Input:} Number of initial samples $N_0$, number of ongoing samples $N_t$, number of policies in the ensemble $M$, number of time steps $T$. \\
    \textbf{Initialize:} Initial World Model $\hat{f}_0$ comprised of $M$ models. \; \\
    Collect $N_0$ samples with a random policy and initialize data set $\mathcal{D}_0 = \{(s_t, a_t), s_{t+1}\}_{t=1}^{N_0}$. \; \\
    \For{$t= 1, \ldots, T-1$}{
      1. Train $\hat{f}_{t-1}$ with $\mathcal{D}_t$ to derive $\hat{f}_{t}$. \;\\
      2. Select $\lambda$ using Algorithm \ref{Alg:ol}. \; \\
      3. Train exploration policy $\pi_{\phi_t}$ using Equation \ref{eqn:explore} in $\hat{f}_{t}$. \; \\
      4. Collect new samples $\mathcal{D}_{new} = \{(s_t, a_t), s_{t+1}\}_{t=1}^{N_t}$ in the environment with $\pi_{\phi_t}$, where $N_t$ is defined as the number of time steps required for Algorithm \ref{Alg:earlystop} to return. \; \\
      5. $\mathcal{D}_{t+1} \leftarrow \mathcal{D}_t \cup \mathcal{D}_{new}$ \; \\
      6. Set $\alpha_t = \mathcal{L}(\hat{f}_t(\mathcal{D}_{new}))$
     }
     \caption{RP1: Ready Policy One}
    \label{Alg:rp1}
    \end{algorithm}
\end{minipage}
}

We now present our algorithm: Ready Policy One (RP1). At each iteration we select a policy objective ($\lambda$) to maximize sample utility, and train a policy on this objective. The policies are trained using the original PPO \cite{schulman2017proximal} loss function, but we use the training approach in \cite{schulman2015trust} as this combination delivered more robust policy updates.\footnote{Full implementation details can be found in Appendix \ref{sec:implementation}.}

Once the policy has been trained inside the model, we use it to generate samples in the real environment. These samples continue until our early stopping mechanism is triggered, and we have sufficiently diverse data to retrain the model. The full procedure is outlined in Algorithm \ref{Alg:rp1}.

The overall aim is to therefore determine which part of the model space is both high value and unknown, so that our trained sampling policy can obtain enough data samples pertaining to those regions of the environment.

%% file: theory.tex




%% file: experiments.tex
\section{Experiments}
\label{sec:experiments}

The primary goal of our experiments it to evaluate whether our active learning approach for MBRL is more sample efficient than existing approaches. In particular, we test RP1 on a variety of continuous control tasks from the $\mathrm{OpenAI}$ $\mathrm{Gym}$ \cite{gym}, namely: $\mathrm{HalfCheetah}$, $\mathrm{Ant}$, $\mathrm{Swimmer}$ and $\mathrm{Hopper}$, which are commonly used to test MBRL algorithms. For specifics, see Appendix \ref{sec:implementation}. In order to produce robust results, we run all experiments for ten random seeds, more than typically used for similar analyses \cite{deeprlmatters}. 

Rather than individual algorithms, we compare against the two approaches most commonly used in MBRL: 
\begin{itemize}
    \item \textbf{Greedy}: We train the policy to maximize reward in the model, and subsequently add noise to discover previously unseen states. This is the approach used in ME-TRPO \cite{ME-TRPO}.
    \item \textbf{Variance + Reward (V+R)}: We train the policy with $\lambda=0.5$, producing a fixed degree of priority for reward and model entropy. This resembles methods with exploration bonuses such as \cite{vime}.
\end{itemize}

We note that these baselines are non-trivial. In particular, ME-TRPO is competitive with state of the art in MBRL. In fact, for two of the tasks considered ($\mathrm{Swimmer}$ and $\mathrm{Hopper}$) it outperformed all other approaches in a recent paper benchmarking MRBL methods \cite{benchmarkingmbrl}.

We also compare against the same policy gradients algorithm as a model free baseline, which we train for $10^6$, $5\times 10^6$ and $10^7$ timesteps. This provides an indication of the asymptotic performance of our policy, if trained in the true environment.

\begin{table}[ht]
\small
    \centering
    \caption{\small{Median best performance at a given timestep for ten seeds. Bold indicates the best performing algoirthm. T1 corresponds to the t-stat for RP1 vs. Greedy, T2 corresponds to the t-stat for RP1 vs. V+R. * indicates $p < 0.05$.}}
    \scalebox{0.9}{
    \begin{tabular}{ccccc|cc}
    \toprule
    & Timesteps & Greedy & V+R &  RP1 & T1 & T2 \\
    \midrule
    $\mathrm{HalfCheetah}$ & $10^4$ & -0.95 & -1.1 & \textbf{100.51} & 5.39* & 4.34* \\
    $\mathrm{Ant}$ & $10^4$ & 94.72 & 95.07 & \textbf{113.63} & 4.02* & 3.08* \\
    $\mathrm{Swimmer}$ & $10^4$ & 1.08 & 1.07 & \textbf{3.24} & 1.2 & 1.98 \\
    $\mathrm{Hopper}$ & $10^4$ & 76.5 & 139.03 & \textbf{322.22} & 4.32* & 3.42* \\
    \midrule
    $\mathrm{HalfCheetah}$ & $10^5$ & 260.92 & 283.27 & \textbf{390.49} & 3.89* & 2.62* \\
    $\mathrm{Ant}$ & $10^5$ & 186.36 & 217.08 & \textbf{238.4} & 3.83* & 3.11* \\
    $\mathrm{Swimmer}$ & $10^5$ & 61.76 & 62.89 & \textbf{64.19} & -0.11 & -1.09 \\
    $\mathrm{Hopper}$ & $10^5$ & 487.25 & 570.09 & \textbf{619.73} & 3.52* & 2.21 \\   
    \bottomrule
    \end{tabular}}
\label{table:main}
\end{table}

\begin{figure}[h]
    \centering\includegraphics[width=.65\linewidth]{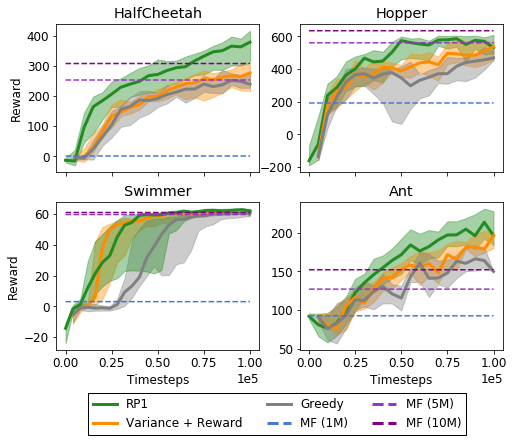}
    \vspace{-3mm}
    \caption{\small{Median performance across $10$ seeds. Shaded regions correspond to the Inter-Quartile Range.}}
    \label{figure:main}
\end{figure}

Table \ref{table:main} and Fig \ref{figure:main} show the main results, where RP1 outperforms both the greedy baseline and the fixed variance maximizing (V+R) approach. Furthermore, we perform Welch's unequal variances t-test and see that in most cases the results are statistically significant, aside from $\mathrm{Swimmer}$ at $10^5$ timesteps where all three methods have converged to the optimal solution. In addition, we observe that RP1 is able to achieve strong performance vs. model-free in fewer timesteps than the existing baselines. 

Interestingly, we see that simply adding a fixed entropy term (V+R) into the reward function gives improved performance over the baseline greedy approach. This corroborates with findings in \cite{ObsDropout}, where there is, in most tasks, a correlation between regions that are difficult to predict and regions of high reward. However our findings also suggest that this is not always the case, and having the ability to adjust how much we focus on such regions of disagreement is vital. We hypothesize that the fixed V+R approach may collect too many high-variance samples for certain tasks, since we do not tune $\lambda$, nor limit batch size. As a result, the trajectories gathered do not necessarily result in strong policy performance, unlike RP1, which aims to maximize policy performance through the data collected.

\begin{figure}[h]
    \centering\includegraphics[width=.7\linewidth]{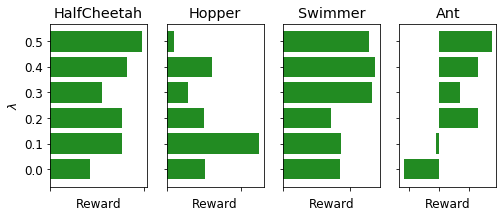} 
    \vspace{-5mm}
    \caption{\small{Mean one-step policy improvement after a given $\lambda$ value, for all ten seeds of RP1.}}
    \label{figure:lambda_analysis}
\end{figure}

We support this hypothesis with Fig. \ref{figure:lambda_analysis} where, we show the normalized change in reward for different $\lambda$ values in each task. In particular, we observe for $\mathrm{HalfCheetah}$, $\mathrm{Swimmer}$ and $\mathrm{Ant}$, a greater focus on uncertainty appears positively related to faster learning. However, the opposite is true for $\mathrm{Hopper}$. The benefit of our mechanism is the ability to dynamically learn this preference, and thus adapt to the current optimization landscape.

\begin{figure}[h]
    \centering\includegraphics[width=.65\linewidth]{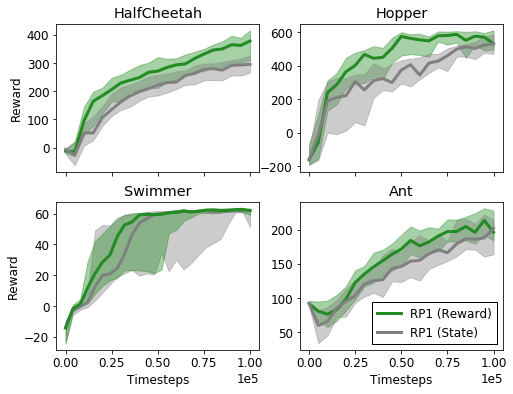} 
    \vspace{-3mm}
    \caption{\small{Median performance across $10$ seeds. Shaded regions correspond to the Inter-Quartile Range.}}
    \label{figure:ablation_statevar}
\end{figure}

Next we study the choice of model variance used in the reward function. Other work, such as \cite{pathak} use the variance over the next state prediction, whereas RP1 uses the variance over the reward. Fig \ref{figure:ablation_statevar} and Table \ref{table:ablation_statevar} show that next state variance is a less effective approach. This is likely due to over-emphasis of regions of the state space that are inherently hard to predict, but do not impact the ability to solve the task \cite{schmidhuber_fun}. 

\vspace{-4mm}
\begin{table}[h]
\small
    \caption{\small{Study for the choice of error to maximize. Results show the median best performance at $10^5$ timesteps for ten seeds. The highest performing value for each environment is bolded.}}
    \centering
    \scalebox{0.85}{
    \begin{tabular}{l*{5}{c}r}
    \toprule
    & $\mathrm{HalfCheetah}$ & $\mathrm{Ant}$ & $\mathrm{Swimmer}$ & $\mathrm{Hopper}$  \\
    \midrule
    State & 319.41 & 214.33 & 63.47 & 549.19 \\
    Reward & \textbf{390.49} & \textbf{238.4} & \textbf{64.19} & \textbf{619.73}  \\
    \bottomrule
    \end{tabular}}
\label{table:ablation_statevar}
\end{table}

Finally, we consider the individual components of the RP1 algorithm. We evaluate two variants: RP1 ($\lambda=0$), where we remove the online learning mechanism and train a greedy policy, and RP1 (No EarlyStop), where we remove the early stopping mechanism and use a fixed batch size. Results are shown in Table \ref{table:ablation_rp1}, and Figs \ref{fig:abl_pca} and \ref{fig:abl_adapt} in Appendix \ref{sec:ablations}. 

\vspace{-3mm}
\begin{table}[ht]
\small
    \centering
    \caption{\small{Ablation study for the key components of RP1. Results show the median performance at $10^5$ timesteps for $10$ seeds. The highest performing value for each environment is bolded.}}
    \scalebox{0.9}{
    \begin{tabular}{l*{5}{c}r}
    \toprule
    & $\mathrm{HalfCheetah}$ & $\mathrm{Ant}$ & $\mathrm{Swimmer}$ & $\mathrm{Hopper}$  \\
    \midrule
    Baseline & 283.27 & 186.36 & 62.89 & 487.25 \\
    RP1 ($\lambda=0$) & 319.14 & 223.21 & 41.14 & 603.52 \\
    RP1 (No EarlyStop) & 247.82 & 197.95 & 41.04 & 595.77 \\
    RP1 & \textbf{390.49} & \textbf{238.4} & \textbf{64.19} & \textbf{619.73} \\
    \bottomrule
    \end{tabular}}
\label{table:ablation_rp1}
\end{table}

We observe that the improvements attributed to RP1 are not down to any single design choice, and the individual components complement each other to provide significant overall gains. For example, by conducting purely noise-based exploration ($\lambda =0$), we lose the flexibility to target specific regions of the state-space. On the other hand, by removing our early stopping mechanism (No EarlyStop), we acquire a trajectory dataset for our model that has too much redundant data, reducing sample efficiency. Nonetheless, we believe adding either of these components to existing MBRL methods, which either have a fixed temperature parameter ($\lambda$) or fixed data collection batch size, would lead to performance gains.

%% file: conclusions.tex
\section{Conclusion and Future Work}
\label{sec:conclusion}

We presented Ready Policy One (RP1), a new approach for Model-Based Reinforcement Learning (MBRL). RP1 casts data collection in MBRL as an active learning problem, and subsequently seeks to acquire the most informative data via an exploration policy. Leveraging online learning techniques, the objective function for this policy adapts during optimization, allowing RP1 to vary its focus on the often fruitful reward function. We showed in a variety of experiments that RP1 significantly increases sample efficiency in MBRL, and we believe it can lead to new state of the art when combined with the latest architectures. 

We are particularly excited by the many future directions from this work. Most obviously, since our method is orthogonal to other recent advances in MBRL, RP1 could be combined with state of the art probabilistic architectures \cite{pets}, or variational autoencoder based models \cite{worldmodels, planet}. 

In addition, we could take a hierarchical approach, by ensuring our exploration policies maintain core behaviors but maximize entropy in some distant unexplored region. This would require behavioral representations, and some notion of distance in behavioral space \cite{CoReyes2018SelfConsistentTA}, and may lead to increased sample efficiency as we could better target specific state action pairs.

%% file: appendix.tex
\appendix
\onecolumn
\section{Additional Ablation Studies}\label{sec:ablations}

In this section we show the full results from the ablation study in Table \ref{table:ablation_rp1}.
\vspace{-5mm}
\begin{figure}[H]
    \begin{minipage}{0.99\textwidth}
    \centering\subfigure[\textbf{HalfCheetah}]{\includegraphics[width=.24\linewidth]{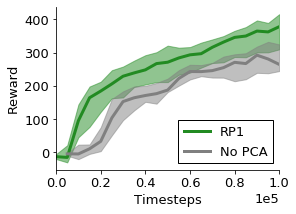}}
    \centering\subfigure[\textbf{Ant}]{\includegraphics[width=.24\linewidth]{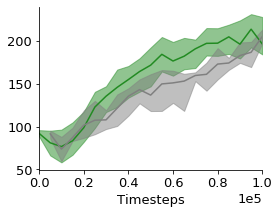}} 
    \centering\subfigure[\textbf{Hopper}]{\includegraphics[width=.24\linewidth]{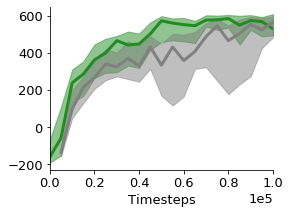}}
    \centering\subfigure[\textbf{Swimmer}]{\includegraphics[width=.24\linewidth]{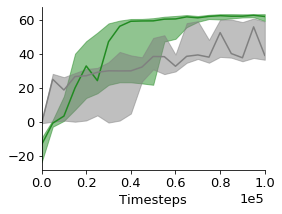}}
    \vspace{-3mm}
    \end{minipage}
    \caption{\small{Ablation study where we consider removing the early stopping mechanism. All results show the median performance across ten seeds.}}
    \label{fig:abl_pca}
\end{figure}
\vspace{-3mm}
\begin{figure}[H]
    \begin{minipage}{0.99\textwidth}
    \centering\subfigure[\textbf{HalfCheetah}]{\includegraphics[width=.24\linewidth]{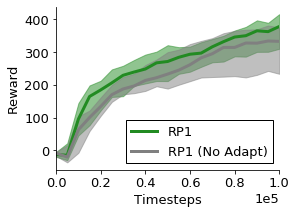}}
    \centering\subfigure[\textbf{Ant}]{\includegraphics[width=.24\linewidth]{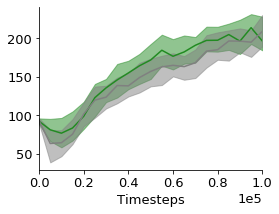}} 
    \centering\subfigure[\textbf{Hopper}]{\includegraphics[width=.24\linewidth]{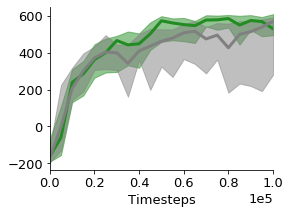}}
    \centering\subfigure[\textbf{Swimmer}]{\includegraphics[width=.24\linewidth]{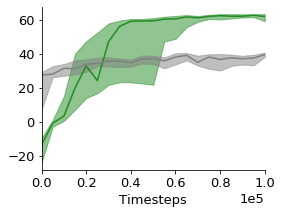}}
    \vspace{-3mm}
    \end{minipage}
    \caption{\small{Ablation study where we consider removing the adaptive mechanism. All results show the median performance across ten seeds.}}
    \label{fig:abl_adapt}
\end{figure}

\section{Implementation Details}\label{sec:implementation}
In terms of approach, we follow ME-TRPO \cite{ME-TRPO} with some adjustments. Instead of using TRPO loss \cite{schulman2015trust}, we leverage the first-order approximation loss in PPO \cite{schulman2017proximal}. We do not apply parameter noise, and we modify policy training slightly by ensuring at least 10 updates are performed before termination is considered, which we find helps to improve convergence. We do not apply GAE nor the overall training approach in \cite{schulman2017proximal}, as this introduced instabilities. We instead found that generating large batch sizes gave better and more consistent performance, which corroborates the findings in \cite{aredeeppgs} with respect to true policy gradients.

We augment each environment with an additional state which contains velocity information. This is also done in the `FixedSwimmer' environment in \cite{benchmarkingmbrl}, and allows us to infer the reward from the states directly. It must also be noted that in the original `rllab' \cite{rllab} environments used in \cite{ME-TRPO}, one of the observable states was the velocity state used to calculate rewards, and we therefore mirror this in our OpenAI Gym \cite{gym} implementation; we do not anticipate there to be any problem integrating reward prediction with our framework. Furthermore, we provide this state in both the model-free and model-based benchmarks to ensure there is no advantage, and do not notice any noticeable improvement in the model-free setting when this is provided; we hypothesize that some close proxy to the true velocity state already exists in the original state-space. We remove contact information from all environments, and instead of `Swimmer-v2' we use the aforementioned `FixedSwimmer', since this can be solved by our policy in a model-free regime. We remove early stopping from Hopper since we found it was necessary for convergence, but left the early stopping in for Ant since it was possible to train performant policies. We train the policy for 100 time steps in HalfCheetah and Ant, and for 200 time steps in Swimmer and Hopper. In experiments without the early stopping mechanism, data collection defaults to 3,000 timesteps per iteration. Full hyperparameter values can be found in Table \ref{table:hyperparameters}.

We use the following approach to normalize $G$ to produce $\hat{G}$; for convenience, we write $\hat{G}_{\phi_{t}}(\theta_{t+1})$ as $\hat{G}_t$.
\begin{align}
    \hat{G}_t = \frac{{G}_t - \frac{1}{5}\sum_{\tau=t-5}^{t-1}(G_{\tau})}{l_{\mathrm{val}}}
\end{align}
where $l_\mathrm{val}$ is the final model validation loss from the iteration $t-1$.

Attention should be drawn to the $\alpha$ parameter used to determine early stopping in Algorithm \ref{Alg:earlystop}. For the tasks we test on, we choose to fix this to $0.0005$, and therefore do not tune it to be task specific. We found that at this setting of $\alpha$, the early stopping mechanism generally collects significantly fewer than the default 3,000 samples (which acts as an upper bound in RP1), but can still expand the batch size collected to the full amount where appropriate (i.e., under policies that provide non-homogenous trajectories).

\begin{table}[h]
\small
    \caption{\small{Hyperparameters used in the Policy}}
    \centering
    \scalebox{0.9}{
    \begin{tabular}[t]{{c c}}
    \toprule
    Hyperparameter Name & Value  \\
    \midrule
    Optimizer & Adam \\
    Learning Rate & 3e-4 \\
    Loss & PPO \\
    Discount Factor & 0.99 \\
    Batch Size & 50,000 \\
    Epochs per Batch & 10 \\
    $\epsilon$-clip & 0.2 \\
    Default Action $\sigma$ & 0.5 \\
    Hidden Layer Size & 32 \\
    Number of Hidden Layers & 2 \\
    Activation Function & ReLU \\
    \bottomrule
    \end{tabular}}
    \end{table}
    \begin{table}[h]
    \caption{\small{Hyperparameters used in the World Model}}
    \centering
    \scalebox{0.9}{
    \begin{tabular}[t]{{c c}}
    \toprule
    Hyperparameter Name & Value  \\
    \midrule
    Optimizer & Adam \\
    Learning Rate & 1e-3 \\
    Train/Validation Split & 2:1 \\
    Number of Models & 5\\
    Batch Size & 1,024 \\
    Hidden Layer Size & 1,024\\
    Number of Hidden Layers & 2\\
    Activation Function & ReLU\\
    Early Stopping $\alpha$ in PCA & 0.0005\\
    \bottomrule
    \end{tabular}}

\label{table:hyperparameters}
\end{table}